\documentclass{article}
\usepackage{ijcai17}

\usepackage{times}

\usepackage{amsfonts}
\usepackage{hyperref}
\usepackage{graphicx}

\title{Beating the World's Best at Super Smash Bros. Melee with Deep Reinforcement Learning}

\author{
    Vlad Firoiu\\
    MIT\\
    vladfi1@mit.edu
  \And
    William F. Whitney\\
    NYU\\
    wwhitney@cs.nyu.edu
    \And
    Joshua B. Tenenbaum\\
    MIT\\
    jbt@mit.edu
}

\begin{document}

\maketitle

\begin{abstract}
There has been a recent explosion in the capabilities of game-playing artificial intelligence. Many classes of RL tasks, from Atari games to motor control to board games, are now solvable by fairly generic algorithms, based on deep learning, that learn to play from experience with minimal knowledge of the specific domain of interest. In this work, we will investigate the performance of these methods on Super Smash Bros. Melee (SSBM), a popular console fighting game. The SSBM environment has complex dynamics and partial observability, making it challenging for human and machine alike. The multi-player aspect poses an additional challenge, as the vast majority of recent advances in RL have focused on single-agent environments. Nonetheless, we will show that it is possible to train agents that are competitive against and even surpass human professionals, a new result for the multi-player video game setting.
\end{abstract}

\section{Introduction}

The past few years have seen a renaissance of sorts for neural network models in AI and machine learning. Driven in part by hardware advances in the GPUs that accelerate their training, the first breakthroughs came in 2012 when convolutional architectures were able to achieve record performance on image classification \cite{krizhevsky2012imagenet} \cite{1202.2745}. Today the technique is known as \emph{Deep Learning} due to its use of many layers that build up increasingly abstract representations from raw inputs.

In this paper we focus not on vision but on game-playing. As far back as the early 90's, neural networks were used to reach expert-level play on Backgammon \cite{tdGammon}. More recently, there have been breakthroughs on learning to play various video games \cite{mnih-atari-2013}. Even the ancient board game Go, which for long has thwarted attempts by AI researchers to build human-level programs, fell to a combination of neural networks and Monte-Carlo Tree Search \cite{alphago}.

\section{The SSBM Environment}

We focus on Super Smash Bros. Melee (SSBM), a fast-paced multi-player fighting game released in 2001 for the Nintendo Gamecube. SSBM has steadily grown in popularity over its 15-year history, and today sports an active tournament and professional scene. The metagame is constantly evolving as new mechanics are discovered and refined and top players push each other to ever greater levels of skill.

From an RL standpoint, the SSBM environment poses several challenges - large and only partially observable state, complex transition dynamics, and delayed rewards. There is also a great deal of diversity in the environment, with 26 unique characters and a multitude of different stages. The partial observability comes from the limits of human reaction time along with several frames of built-in input delay, which forces players to anticipate their opponent's actions ahead of time. Furthermore, being a multi-player game adds an entirely new dimension of complexity - success is no longer a single, absolute measure given by the environment, but instead must be defined relative to a variable, unpredictable adversary.

\begin{figure}[h]
\centering
\includegraphics[width=0.4 \textwidth]{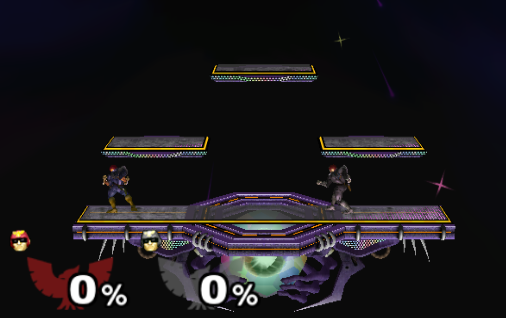}
\caption{The Battlefield Stage}
\end{figure}

\subsection{State, Action, Reward}

Many previous applications of deep RL to video games have used raw pixels as observations. Partly for pragmatic reasons, we instead use features read from the game's memory on each frame, consisting of each player's position, velocity, and action state, along with several other values. This allows us to focus purely on the RL challenge of playing SSBM rather than the perception. In any case, the game features are readily inferred from the pixels; as deep networks are known to perform quite well on vision tasks, we have good reason to believe that pixel-based models would perform similarly. Pixel-based networks would also be better able to deal with projectiles, which we do not currently know how to read from the game memory.

The game runs natively at 60 frames per second, which we lower to 30 by skipping every other frame. No actions are sent on the skipped frames, which is equivalent to the controller not changing state. To better match human play, we would lower this further by skipping more frames, but that would make it impossible to perform certain important actions which humans perform regularly (for example, some characters require releasing the the jump button at most 2 frames after pressing it in order to perform a ``short hop'' instead of the full jump).

The GameCube controller has two analog sticks, five buttons, two triggers, and a directional pad, all of which are relevant in SSBM. To make things easier, we eliminate most of the inputs, leaving only 9 discrete positions on the main analog stick and 5 buttons (at most one of which may be pressed at a time), for a total of 54 discrete actions. This suffices for the majority of the relevant actions in SSBM, although proficient humans routinely make use of controller inputs outside this limited set (such as precise angles and partial tilts of the control stick).


The goal of SSBM is to knock out (KO) the opponent by sending them out of bounds, and we give scores of $\pm 1$ for these events. How far opponents are sent flying when hit depends on their damage, which is displayed on screen. We add the damage dealt (and subtract the damage taken) from the score, with a small weighting factor. Although not the ultimate objective, this reward signal is very important to humans, so we felt it was appropriate to include it. Without it, learning from the very sparse KO signal alone would be very difficult.

Players re-spawn in the middle of the stage after being KOed. In tournaments, games are won after four KOs. To simplify navigating through the SSBM menus we instead set the game mode to infinite time and arbitrarily mark off episodes every few seconds.

\section{Methods}

We used two main classes of model-free RL algorithms: Q-learning and policy gradients. While standard, we follow with a brief review of these techniques. Henceforth, we will use $s$ to denote states, $a$ to denote actions, and $r$ to denote rewards, all three of which may be optionally indexed by a time step. Capital letters denote random variables.

\subsection{Q-learning}

In Q-learning, one attempts to learn a function mapping state-action pairs to expected future rewards:

\begin{equation}
Q_\pi(s_t, a_t) = \mathbb{E}[R_t + \lambda R_{t+1}+\lambda^2 R_{t+2} + \cdots]
\label{eq:Q}
\end{equation}

We assume that all future actions (upon which the $R_i$ are implicitly dependent) are taken according to the policy $\pi$. In practice, we estimate the RHS from a single sampled trajectory, and also truncate the sum in order to reduce the variance of the estimate (at the cost of introducing bias). Our objective function becomes:
\begin{equation}
L = \left( Q(s_t, a_t) - \left[ r_t + \lambda r_{t+1} + \cdots + \lambda^n Q(s_{t+n}, a_{t+n}) \right] \right)^2
\label{eq:qLoss}
\end{equation}
With $Q$ approximated by neural network, we use (batched) stochastic gradient descent on $L$ to learn the parameters. Note that the second (subtracted) $Q$ in the objective is considered a constant with regards to gradients; we wish to adjust $Q$ to become a better predictor of future rewards, not to adjust the future rewards to match the past prediction.

Once we learn $Q_\pi$ for some policy $\pi$, we can construct a new (better) policy $\pi'$ which always takes the best action under the learned $Q_\pi$, and repeat. This is known as policy iteration, and is guaranteed to quickly converge to the optimal policy for small environments. Of course, more interesting environments like SSBM have large (and continuous) state spaces, and so it is prohibitive to exhaustively explore the entire space. In such cases it is common to generate experiences using an $\epsilon$-greedy strategy, in which a random action is taken with probability $\epsilon$. To further explore promising actions, we also take actions from a Boltzmann distribution over their predicted $Q$-values. That is, in state $s$ we take action $a$ with probability proportional to $\exp(\tau Q(s, a))$, where $\tau$ is an (inverse) temperature parameter that must be chosen to match the scale of the $Q$-values.

In the RL literature, our approach might be referred to as $n$-step SARSA. DeepMind's original work using deep $Q$-networks (abbreviated \emph{DQN}) on Atari games employed a slightly different algorithm based on the Bellman equation \cite{mnih-atari-2013}:
\begin{equation}
Q_{\hat{\pi}}(s_t, a_t) = \mathbb{E}[R_t + \lambda \max_a Q_{\hat{\pi}}(S_{t+1}, a)]
\label{eq:bellman}
\end{equation}

In principle this would allow one to directly learn $Q$ for the optimal policy $\hat{\pi}$, independent of the policy used to generate the experiences. However, we found this to be much less stable than SARSA, with the $Q$-values rapidly diverging from reality, likely due to the iteration of the maximum operator during training. There exist techniques such as the double-DQN \cite{doubleDQN} to alleviate this effect, which warrant further exploration.

A note about implementation: our $Q$-network does not actually take the action as an input, but instead outputs a vector of $Q$-values for all the actions.

\subsection{Policy Gradient Methods}

Policy gradient methods work slightly differently from $Q$-learning. Their main feature is an explicit representation of the policy $\pi$, which maps states to (distributions over) actions, and which is directly updated based on experience. The REINFORCE \cite{williams1992simple} learning rule is the prototypical example:

\begin{equation}
\Delta \theta = \alpha (R-b) \nabla_\theta \log \pi_\theta(s, a)
\label{eq:REINFORCE}
\end{equation}
Here $R$ is the sampled future reward (possibly truncated, as above), $b$ is a baseline reward, and $\alpha$ is the learning rate. Intuitively, this increases the probability of taking actions that performed better than the baseline, and vice-versa. It can be shown that, in expectation, $\Delta\theta$ maximizes the expected discounted rewards, averaged over all states.


The Actor-Critic algorithm is an extension of REINFORCE that replaces the baseline $b$ with a parameterized function of the state, known as the critic. This critic $V_\pi(s)$ attempts to predict the expected future reward from a state $s$ assuming that the policy $\pi$ is followed, very similar to the above $Q$ function:

\begin{equation}
V_\pi(s_t) = E[R_t + \lambda R_{t+1}+\lambda^2 R_{t+2} + \cdots]
\label{eq:V}
\end{equation}

Ideally, this removes all state-dependent variance from the reward signal, leaving only the action-dependent component or \emph{advantage}, $A(s, a) = Q(s, a) - V(s)$, to inform policy updates. In our experience the value networks perform quite well, explaining about 90\% of the variance in rewards.

One issue that Actor-Critics face is premature convergence to a suboptimal deterministic policy. This is bad because, once the policy is deterministic, different actions are no longer explored, so we never receive evidence that other actions might be better, and so the policy never changes. A simple workaround is to add some $\epsilon$ noise to the policy, like in $Q$-learning. However, because we do not have $Q$-values, we can't explicitly explore similarly-valued actions with similar probabilities. Instead, we add an entropy term to the learning rule (\ref{eq:REINFORCE}) that nudges the policy towards randomness, and we tune the scale $h$ of this entropy term so that the actor neither plunges into deterministism (0 entropy) nor remains stuck at uniform randomness (maximum entropy). Since entropy is simply expected (negative) log-probability, our resulting Actor-Critic policy gradient is:

\[
\Delta \theta = \alpha (A(s, a) - h) \nabla_\theta \log \pi_\theta(s, a)
\]

In this form, we see that the entropy scale $h$ is constant negative distortion on the reward signal. Therefore, like the REINFORCE baseline $b$, $h$ does not affect the overall validity of the policy gradient as a maximizer (in expectation) of total discounted reward. 

Overall our approach most closely resembles DeepMind's Asynchronous Advantage Actor-Critic \cite{a3c}, although we do not perform asynchronous gradient updates (merely asynchronous experience generation). Similar to the $Q$ network, the actor network outputs a vector containing the probabilities of each action.

\subsection{Training}

Despite being 15 years old, SSBM is not trivial to emulate \footnote{We used the dolphin emulator. http://dolphin-emu.org}. Empirically, we found that, while a modern CPU can reach framerates of about 5x real time, those typically found on servers can only manage 1-2x. This is quite slow compared to the performance-engineered Atari Learning Environment, which can run Atari games over one hundred times faster than real time. This means that generating experiences (state-action-reward sequences) is a major bottleneck. We remedy this by running many different emulators in parallel, typically 50 or more per experiment. \footnote{Computing resources were provided by the Mass. Green High-Performance Computing Center.}

The many parallel agents periodically send their experiences to a trainer, which maintains a circular queue of the most recent experiences. With the help of a GPU, the trainer continually performs (minibatched) stochastic gradient descent on its set of experiences while periodically saving snapshots of the neural network weights for the agents to load. This asynchronous setup technically breaks the assumption of the REINFORCE learning rule that the data is generated from the current policy network (in reality the network has since been updated by a few gradient steps), but in practice this does not appear to be a problem, likely because the gradient steps are sufficiently small to not change the policy significantly in the time that an experience sits in the queue. The upside is that no time is wasted waiting on the part of either the agents or the trainer.

\subsubsection{Hyper-Parameters}

All of our policies used an epsilon value of 0.02. Our discount factor $\lambda$ was set such that rewards 2 seconds into the future were worth half as much as rewards in the present. We tried different values of $n$ in the discounted reward summation and settled on $n=10$.

All of our neural networks (Q, actor, and critic) used architectures with two fully-connected hidden layers of size 128. While far from thorough, our attempts with different architectures did not yield improvements - some, such the 3 x 128 policy network, actually did worse. On the other hand, the number and sizes of the critic layers did not have much of an effect.

Our weight variables were initialized to have random columns of norm 1, and the biases as zero-mean normals with standard deviation 0.1. Our nonlinearity was a smoothed version of the traditional leaky ReLU which we call ``leaky softplus'' (with $\alpha=0.01$):

\[
f_\alpha(x) = \log(\exp(\alpha x) + \exp(x))
\]

\subsubsection{Learning rate and second-order methods}

Whenever gradient descent is employed, one must worry about choosing the right learning rate. It must not be too large, or the local linearity assumption breaks down and the loss fails to decrease (or even diverges). But if too small, then learning is unnecessarily slow. Ideally, the learning rate would be as large as possible, while still ensuring convergence. Often, some hand-tuning suffices; in our case, a learning rate of 1e-4 gave reasonable results.

A more principled approach is to use higher-order derivatives to adjust the learning rate or even the gradient direction. If the error surface is relatively flat, then we can take a larger step; if it is very curved, then we should take a small step. This incidentally solves another issue with first-order methods: that scaling the loss function (or the rewards) translates into an equivalent scaling of the gradients, which can mean a dramatic change in the learning dynamics, even though the optimization problem is effectively unchanged.

In RL, however, we are optimizing more than just a loss function - we are optimizing a policy through policy iteration. This means that we should care about the change in the policy as well as the change in the loss (and when using REINFORCE there isn't really a ``loss'', only a direction in which to improve the policy). This approach, known as Trust Region Policy Optimization \cite{TRPO}, constrains each gradient step so that the change in policy is bounded. This change is measured by the KL divergence between the old policy and the new policy, averaged over the states in our batch:

\[
D(\pi, \pi') = \frac{1}{|S|}\sum_{s\in S} D_{KL}(\pi(s), \pi'(s))
\]

If the old policy is parameterized by $\theta_0$ which we are changing in the $\Delta\theta$ direction, then a second-order approximation of the change in policy is given by:

\[
D(\pi_{\theta_0}, \pi_{\theta_0 + \Delta\theta}) \approx \frac{1}{2} \Delta\theta^T H(\theta_0) \Delta\theta
\]

Here $H(\theta)$ is the Hessian of $D(\pi_{\theta_0}, \pi_\theta)$. Note that there is no first-order term, since $\theta=\theta_0$ is a global minimum of the policy distance. The direction in which the KL divergence is taken also doesn't matter, as it is locally symmetric.

If the policy gradient direction is $g$, our goal then is to maximize $\Delta\theta^T g$ (the progress made in improving the policy) subject to the constraint

\[
\frac{1}{2} \Delta\theta^T H \Delta\theta \leq c
\]

Here $c$ is a our chosen bound on the change in policy. The method of Lagrange multipliers shows that the optimal direction for $\Delta\theta$ is given by the solution to $Hx = g$ (which we then rescale to satisfy the constraint). Unfortunately, $H$ is in practice too big to invert or even store in memory, as it is quadratic in the number of parameters, which is already quite large for neural networks. We thus resort to the Conjugate Gradient method \cite{cg}, which only requires the ability to take matrix-vector products with $H$. This we can do as follows:

\[
H x = \left[\nabla_\theta \left(x^T \nabla_\theta D(\pi_{\theta_0}, \pi_\theta) \right) \right]_{\theta=\theta_0}
\]

Note that we are only taking gradients of scalars, which can be done efficiently with automatic differentiation. Each step of conjugate gradient descent improves our progress in the direction of the policy gradient $g$ within the constrained policy region, at the cost of extra computation time. In practice, we found that a policy bound of $10^{-6}$ and 10-20 conjugate gradient iterations worked best.

\section{Results}

Unless otherwise stated, all agents, human or AI, played as Captain Falcon on the stage Battlefield \footnote{Considered the best stage for competitive play.}. We chose Captain Falcon because he is one of the most popular characters, and because he doesn't have any projectile attacks (which our state representation lacks). Using only one character and stage greatly simplifies the environment and makes it possible to directly compare learning curves and raw scores.

\subsection{In-game AI}

We began by testing the RL algorithms against the in-game AI. After appropriate parameter tuning, both $Q$ learners and actor-critics proved capable of defeating this AI at its highest difficulty setting, and reached similar average reward levels within a day.


\begin{figure}[h]
\includegraphics[width=0.4\textwidth]{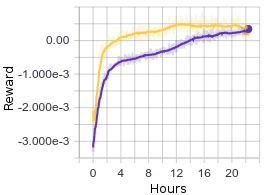}
\caption{Learning curves for Actor-Critic (purple) and ``DQN'' (yellow) against the in-game AI. Y-axis is average reward, X-axis is hours.}
\end{figure}

For each algorithm, we found little variance between experiments with different initializations. However, the two algorithms found qualitatively different policies from each other. Actor-Critics pursued a standard strategy of attacking and counter-attacking, similar to the way humans play. $Q$-learners on the other hand would consistently find the unintuitive strategy of tricking the in-game AI into killing itself. This multi-step tactic is fairly impressive; it involves moving to the edge of the stage and allowing the enemy to attempt a 2-attack string, the first of which hits (resulting in a small negative reward) while the second misses and causes the enemy to suicide (resulting in a large positive reward).

\subsubsection{OpenAI Baseline}

OpenAI has released Gym and Universe, which provide a uniform RL interface to a collection of various environments (such as Atari) \cite{gym}. They also provide a ``starter agent'' implementing the A3C algorithm as a baseline for solving the Gym/Universe RL tasks \footnote{http://github.com/openai/universe-starter-agent}. While our main work does not use this interface, as it lacks support for multi-agent environments, we have implemented SSBM as a Gym environment \footnote{http://github.com/vladfi1/gym-dolphin} (with the in-game AI as the opponent) for easy access. This allowed us to run (a slightly modified version of \footnote{http://github.com/vladfi1/universe-starter-agent}) OpenAI's starter agent on the same task from above: C. Falcon vs max level C. Falcon on Battlefield. However, after running for several days on a 16-core machine, the average reward never surpassed -1e-3, a level which both our DQN and Actor-Critic were able to reach in only a few hours. This suggests that SSBM, even when using the underlying game state instead pixels, is somewhat more difficult than the Atari environments for which the starter agent was built.

\subsection{Self-play}

The agents trained against the in-game AI, while successful, would not pose a challenge to even low-level competitive human players. This is due to the quality of their opponent - the in-game AI pursues a very specific (and, for the $Q$-learner, exploitable) strategy which was does not reflect how experienced players actually play. Without having ever played against a human-level opponent, it is not surprising that the trained agents are themselves below human-level.

By switching the player structs in the state representation, we can have a network play as either player 1 or 2, allowing it to train against old versions of itself in a similar fashion to AlphaGo \cite{alphago}. After a week of self-training an Actor-Critic, our network exhibited very strong play, similar to an expert (if repetitive) human. The author, himself a mid-level player, was hard-pressed to defeat this AI through conventional tactics. After another week of training, we brought a copy of the network to two major tournaments, where it performed favorably against all professional players who were willing to face it.

\begin{table}[h]
  \centering
    \begin{tabular}{llll}
    Opponent & Rank & Kills & Deaths \\
    \hline
    S2J & 16 & 4 & 2 \\
    Zhu & 31 & 4 & 1 \\
    Gravy & 41 & 8 & 5 \\
    Crush & 49 & 3 & 2 \\
    Mafia & 50 & 4 & 3 \\
    Slox & 51 & 6 & 4 \\
    Redd & 59 & 12 & 8 \\
    Darkrain & 61 & 12 & 5 \\
    Smuckers & 64 & 8 & 5 \\
    Kage & 70 & 4 & 1 \\
    \end{tabular}
  \caption{Some results against ranked SSBM players. Rankings from \url{http://wiki.teamliquid.net/smash/SSBM_Rank}. S2J is considered by some to be the best Captain Falcon player in the world.}
\end{table}

Even this very well-trained network exhibited some strange weaknesses, however. One particularly clever player found that the simple strategy of crouching at the edge of the stage caused the network to behave very oddly, refusing to attack and eventually KOing itself by falling off the other side of the stage. One hypothesis to explain this weakness is the lack of diversity in training - since the network only played against old copies of itself, it never encountered such a degenerate strategy.

Another limitation is that this network was only trained to play as and against a specific character and on a particular stage, and predictably performs much worse if these variables are changed. Our attempts to train networks to play as multiple characters at once - that is, to simultaneously train on experiences generated from multiple characters' points of view - did not have much success. Anecdotally, we observed that these networks would not appropriately change their strategy based on their character, choosing to use moves from the rather small intersection of the ``good'' moves of each character. This is somewhat similar to autoencoders that learn to generate blurry images that represent the ``average'' input of their dataset. 

\subsection{Agent Diversity}

The simple solution to playing multiple characters is to use a different network for each character. We did this for the six most popular competitive characters (Fox, Falco, Sheik, Marth, Peach, and Captain Falcon), and had the networks train against each other for several days. The results were fairly good, with the networks becoming challenging for the author to play against. In addition, these did not exhibit the strange behavior of the earlier Falcon-bot. We suspect that this is due to the added uncertainty in the environment from training against different opponents.

This set of six networks then became opponents against which to train future networks, providing a concrete benchmark for measuring performance. Empirically, none of these future attempts were able to find degenerate counter-strategies to the benchmark networks, so we tentatively declare that weakness resolved.


\subsection{Character Transfer}

When training a network to play as a new character, we found it more efficient to initialize from an already-trained network than from scratch. We can measure this in the amount of time taken to reach 0 average reward against the benchmark set of agents.

\begin{table}[h]
  \centering
  	\resizebox{\columnwidth}{!}{%
    \begin{tabular}{l|lllllll}
     & Scratch & Sheik & Marth & Fox & Falco & Peach & Falcon\\
    \hline
    Sheik  & 36 & 0 & 4 & 7 & 7 & 3 & 9\\
    Marth  & 40 & 5 & 0 & 11 & 10 & 7 & 10 \\
    Fox    & 31 & 8 & 6 & 0 & 2 & 6 & 7\\
    Falco  & 35 & 9 & 6 & 2 & 0 & 7 & 5\\
    Peach  & 26 & 2 & 4 & 5 & 5 & 0 & 6 \\
    C. Falcon & 53 & 9 & 11 & 13 & 12 & 10 & 0 \\
    \end{tabular}
    }
    \caption{Transfer times (in hours) for Actor-Critics. We consider a network ``trained'' once it reaches 0 mean reward against the benchmark agents.}
\end{table}

\begin{figure}[h]
\centering
\includegraphics[width=0.4 \textwidth]{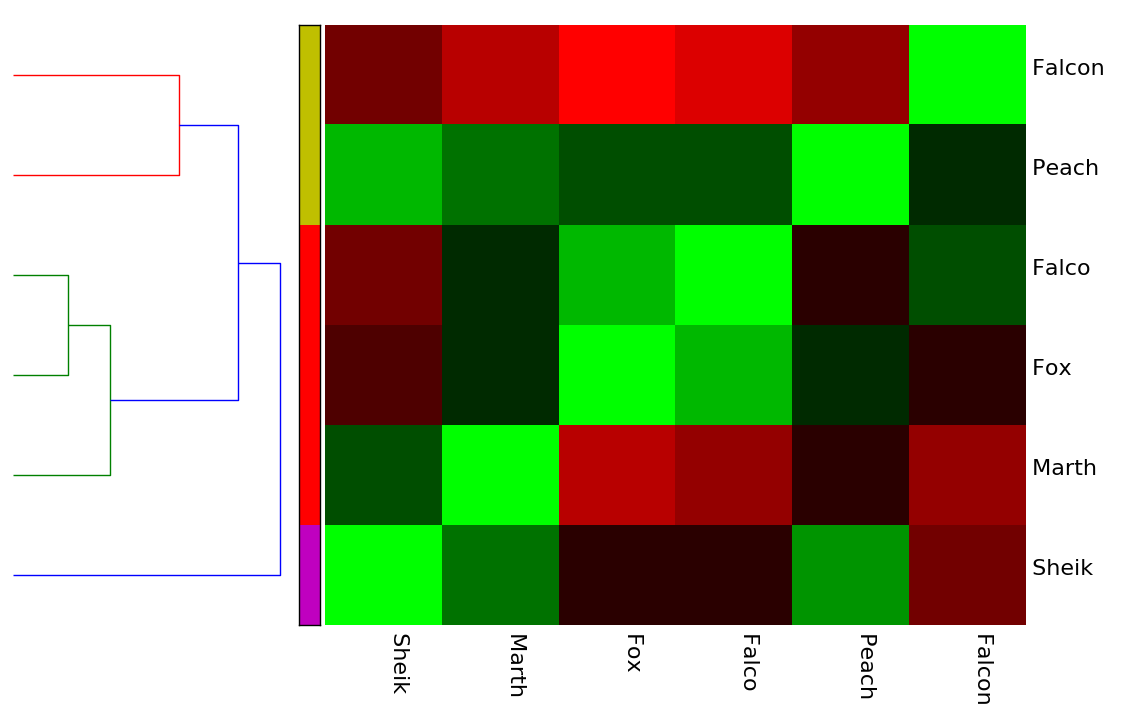}
\caption{Hierarchical clustering of the characters by transfer time. Fox and Falco, considered to be ``clone'' characters, cluster tightly together.}
\end{figure}


By this measure, transfer provides a significant speedup to training. This is especially true for similar pairs of characters, such as Fox and Falco. On the whole these results are unsurprising, as many basic tactics (staying on the stage, attacking in the opponent's direction, dodging or shielding when the opponent attacks) are universal to all characters.

The data also reveal the overall ease of playing each character - Peach, Fox, and Falco all trained fairly quickly, while Captain Falcon was significantly slower than the rest. This to some extent matches the consensus of the SSBM community, which ranks the characters (in decreasing order of strength) as: Fox, Falco, Marth, Sheik, Peach, C. Falcon. The main difference is that Peach performs better than would be naively expected from the community rankings. This is likely due to her very quick and powerful attacks, which are easier for RL agents to learn to use compared to the movement speed offered by other characters like Marth and C. Falcon.


\section{Discussion}

\subsection{Actor-Critic vs Q-Learning}

We found that Q-learners did not perform well when learning from self-play, or in general when playing against other networks that are themselves training. It could be argued that learning the $Q$-function is intrinsically harder than learning a policy. This technically true in the sense that from $Q_\pi$ one can directly get a policy that performs as least as well as $\pi$ (by playing greedily), but from $\pi$ it is not easy to get $Q_\pi$ (that's the entire challenge of Q-learning).

However, we found that Q-learners perform reasonably well against fixed opponents, such as the in-game AI and the set of benchmark networks. This leads us to believe that the issue is the non-stationary nature of playing against agents that are also training. In this scenario the $Q$ function has to keep up with the not only the policy iteration but also the changes in the opponent.

\subsection{Exploration vs Exploitation}

Our main method for quantitatively measuring the tendency of an agent to explore different actions is through the average entropy of its policy. For $Q$-networks, this is directly controlled by the temperature parameter. For Actor-Critics, the entropy scale factor nudges the direction of the policy gradient towards randomness during training.


Looking at only the mean entropy over many states can be misleading, however. Typically the minimum entropy quickly dips below 0.5, while the average remains above 3. In many cases we found these seemingly high-entropy agents to actually play very repetitively. This suggests that on most frames, which action is taken is largely irrelevant to the agent's performance. Indeed, once an attack is initiated in SSBM, it generally cannot be aborted during its duration, which can last on the order of seconds.

A more principled approach to exploration would attempt to quantify the agent's uncertainty, and prefer to explore actions about which the agent is unsure. Even measuring how much an state/action has been explored can be quite difficult - once this is known, bandit algorithms such as UCB may be applied \cite{pseudo-count}.




\subsection{Action Delay}


The main criticism of our agents is that they play with unrealistic reaction speed: 2 frames (33ms), compared to over 200ms for humans. To be fair, Captain Falcon is, of the popular characters, perhaps the worst equipped to take advantage of this reaction speed, with attacks that take many frames to become active (on the order of 15 frames, or 250ms). Many other characters have attacks that become active in half the time, or even immediately (on the very next frame) - this was an additional reason for using C. Falcon initially.

The issue of reaction time highlights a big difference between these neural net-based agents and humans. The neural net is effectively cloned, fed a state, asked for an action, and then destroyed on each frame. While the cloning and destruction don't really take place, this perspective puts the network in stark contrast to people, who have a memory and respond continually to their sensory experiences and internal thoughts.


The closest a neural network can get to this is via recurrence, where the network outputs not only an action but also a memory state, and is fed not only the current game state but also the previous memory. Unfortunately these network are known to be difficult to train \cite{rnn}, and we were not able to train a competent recurrent agent. This avenue certainly warrants further investigation - recurrent networks would be able to deal with any amount of delay and could even in principle handle projectiles, by learning to remember when they were fired and simulating their trajectory in memory.

Instead, to deal with an action delay of $k$ frames, we use a network that takes in the previous $k+1$ frames as input, along with the actions taken on those frames. This was sufficient to train fairly strong agents with delay 2 or 4, but performance dropped off sharply around 6-10 frames. We suspect that the cause for this drop in performance is not simply the handicap given by the delay, but the further separation of actions from rewards, making it harder to tell which actions were really responsible for the already sparse rewards.


To the best of our knowledge, action delay (and human-like play in general) is not an issue that has been addressed by the (deep) RL community, and remains an interesting and challenging open problem.

\section{Conclusions}

The goal of this work is threefold. We introduce a new environment to the reinforcement learning community, Super Smash Bros Melee, a competitive multi-player game which offers a variety of stages and characters. We analyze the difficulties posed by adapting traditional reinforcement learning algorithms, which are typically designed around a stationary Markov decision process, to multi-player games where the adversary may itself learn. Finally, we demonstrate an agent based on deep learning which sets the state of the art in this environment, surpassing the abilities of ten highly-ranked human players.


\bibliographystyle{named}
\bibliography{ijcai17}

\end{document}